\documentclass[twoside,11pt]{article}

\usepackage{jmlr2e}
\usepackage{subcaption}
\usepackage{qtree}
\usepackage{amsmath}
\usepackage{mathtools}

\DeclarePairedDelimiter{\abs}{\lvert}{\rvert}
\DeclarePairedDelimiter{\norm}{\lVert}{\rVert}

\jmlrheading{1}{2022}{1-1}{10/26}{00/00}{Sherman Siu}

\ShortHeadings{Towards Automating Codenames Spymasters with Deep Reinforcement Learning}{Siu}
\firstpageno{1}

\begin{document}

\title{Towards Automating Codenames Spymasters with Deep Reinforcement Learning}

\author{\name Sherman Siu \email s8siu@uwaterloo.ca \\
       \addr Cheriton School of Computer Science \\
       University of Waterloo\\
       Waterloo, ON N2L 3G1, Canada
    %   \AND
    %   \name Michael I.\ Jordan \email jordan@cs.berkeley.edu \\
    %   \addr Division of Computer Science and Department of Statistics\\
    %   University of California\\
    %   Berkeley, CA 94720-1776, USA
    }

\editor{}

\maketitle

% \begin{abstract}%   <- trailing '%' for backward compatibility of .sty file
% This paper describes the mixtures-of-trees model, a probabilistic 
% model for discrete multidimensional domains.  Mixtures-of-trees 
% generalize the probabilistic trees of \citet{chow:68}
% in a different and complementary direction to that of Bayesian networks.
% We present efficient algorithms for learning mixtures-of-trees 
% models in maximum likelihood and Bayesian frameworks. 
% We also discuss additional efficiencies that can be
% obtained when data are ``sparse,'' and we present data 
% structures and algorithms that exploit such sparseness.
% Experimental results demonstrate the performance of the 
% model for both density estimation and classification. 
% We also discuss the sense in which tree-based classifiers
% perform an implicit form of feature selection, and demonstrate
% a resulting insensitivity to irrelevant attributes.
% \end{abstract}

\begin{keywords}
  reinforcement learning, deep learning, board games, natural language understanding, Codenames
\end{keywords}

\section{Introduction}
In artificial intelligence, games are often used as a benchmark for measuring how well artificial intelligence agents can interact with the world. In the recent years, there have been many breakthroughs in AI for single-player games like Atari games \citep{mnih2013playing} and Go \citep{silver2016alphago}, or competitive multi-player games like Starcraft II \citep{vinyals2019alphastar}. However, much less work has been done investigating multi-player co-operative games. Nevertheless, some work has been made in this area, such as for Hanabi~\citep{lerer2020improving}, a co-operative game with imperfect information and online multiplayer Quake III Arena Capture the Flag \citep{jaderberg2019human}. Although traditional multi-agent reinforcement learning (RL) techniques tend to work well when RL agents work with each other, they fail to work well when co-operating with humans \citep{siu-2021-hanabi-fail,bakhtin-2021-no-press-diplomacy-fail}.

For reinforcement learning, text-based games have received comparatively less attention than other types of games and they are an open problem for AI-based research. One of the main challenges with text-based games is the large action space available to agents. Another challenge is the fact that text-based games often require some common sense about how the world works. Thus, text-based games are a good benchmark for natural language understanding \citep{cote2018textworld}. Thus, Codenames is a good benchmark for both human-AI co-operation and text-based reinforcement learning, which are both several important areas of AI research.

\subsection{Overview of Codenames}
\begin{figure}[htbp]
\centering
\begin{subfigure}{.5\textwidth}
  \centering
  \captionsetup{width=.9\linewidth}
  \includegraphics[width=.9\linewidth]{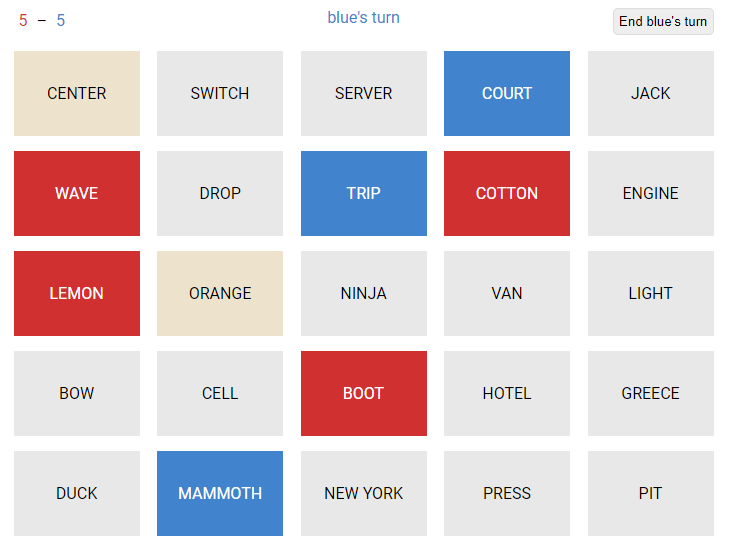}
  \caption{An in-progress Codenames game. Both teams have the same number of cards, but it is blue's turn.}
  \label{fig:sample_game1}
\end{subfigure}%
\begin{subfigure}{.5\textwidth}
  \centering
  \captionsetup{width=.9\linewidth}
  \includegraphics[width=.9\linewidth]{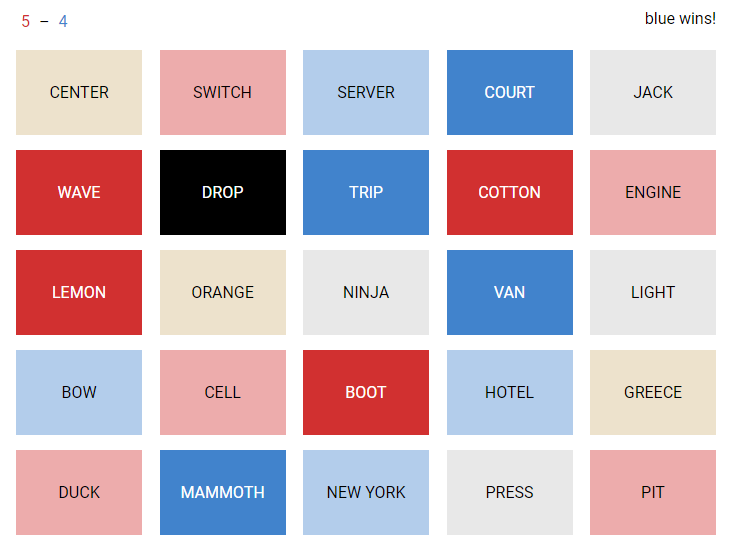}
  \caption{The same game, except the red team accidentally chose the black assassin card. The colours of the remaining cards are revealed.}
  \label{fig:sample_game2}
\end{subfigure}
\caption{A game of Codenames has an associated 5\texttimes5 board of words, each of which has a hidden colour known only to the spymasters of each team until the cards are picked.}
\label{fig:sample_game}
\end{figure}
Codenames is a text-based board game that is both text-based and involves asymmetric cooperation. In Codenames, there are two teams of approximately equal size, a ``red'' team and a ``blue'' team \citep{codenamesrules}. Each team has one spymaster, who has complete information of the board and one or more operatives, who have incomplete information. At the beginning of the game, 25 word cards are drawn from the deck and placed on the board, as shown in Figure \ref{fig:sample_game}. Each word is also assigned one of four labels: red, blue, bystander, and assassin. The goal of each spymaster is to guide their team's operatives to choosing their own team's cards, while avoiding the others. These labels are known to the spymasters from the start of the game, while the operatives know none of the associated labels at the beginning of the game. As the game progresses, the operatives discover the labels of the cards that have been chosen by the operatives of either team.

\begin{figure}[htbp]
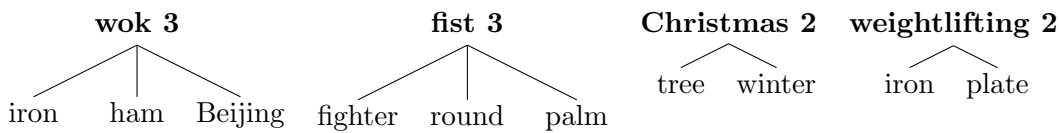

\centering
\Tree[.\textbf{wok 3} iron ham Beijing ]
\Tree[.\textbf{fist 3} fighter round palm ]
\Tree[.\textbf{Christmas 2} tree winter ]
\Tree[.\textbf{weightlifting 2} iron plate ]
\caption{Examples of spymaster hints, with the intended guesser words below. The guesser words are supposed to be similar to the spymaster hints in some way.}
\label{fig:related_words}
\end{figure}

Although the spymaster has perfect information, the spymaster is only allowed to communicate with their team's operatives by providing a single hint each term, consisting of a single word and number, as shown in Figure \ref{fig:related_words} \citep{codenamesrules}. The word cannot be a superstring or substring of any of the words on the board. The word cannot give a directional clue, such as ``left'' to suggest that the correct words are on the left side of the board. The number indicates how many words on the board relate to the hint word. Thus, the spymaster must only provide a hint that relies on the meaning of the words.
% This provides an additional layer of complexity for AI, as modern context-dependent word embeddings, such as from BERT~\citep{devlin-etal-2019-bert} require some sort of context for the embeddings to be context-dependent.

The hint number can also be 0 to indicate that none of the team's remaining words relate to that hint word or infinity to indicate that the team can make as many guesses as they want before choosing to end their turn \citep{codenamesrules}.

Codenames is a zero-sum, turn-based game. Regardless of the team's colour, the first team to start playing has 9 cards belonging to their team and the second has 8 \citep{codenamesrules}. There are always 7 bystander cards and 1 assassin card. After hearing the spymaster's hint, the operatives of each team have the choice of either flipping over a word card or ending their turn. If the card's colour isn't their team's colour, their team's turn immediately ends. Otherwise, the team has the choice to either continue flipping cards or ending their turn. If a team flips the assassin card, that team immediately loses. If none of the teams flip over the assassin card, the first team to flip over all of their cards wins. Hence, spymasters must be careful to not suggest hints that might help the other team or cause the assassin card to be flipped.

\subsection{My Contributions}
% TODO: Change to we/our/us when actually publishing.
To the best of my knowledge, there are no reinforcement learning-based spymaster agents that have been used for Codenames, despite the success of reinforcement learning \citep{mnih2013playing,christiano2017deep-rhlf,silver2018general-alphazero}, even when applied to environments in which natural language processing plays a huge part \citep{ouyang2022training-instructgpt,bakhtin-etal-2022-cicero}.

\begin{enumerate}
    \item As a result, I formulate a game of Codenames from a Spymaster's perspective as a Markov Decision Process (MDP) and implement several OpenAI Gym \citep{brockman2016openai-gym} environments for a spymaster agent and
    \item demonstrate that several agents (SAC, SAC with HER), PPO, and A2C) do not cause converge when trained on all of the above environments. Moreover, both a random policy and a simply greedy policy beat all of the above models on the Codenames environments.
    \item I formulate two simplified versions of the environment called ``ClickPixel'' and  ``Whack-a-Mole'' and demonstrate that while A2C and sometimes PPO converges when the environment is small, they fail to converge as the state space and action space are scaled up.
    % \item I also identify a promising future research direction, where self-supervised pre-training, supervised training, and reinforcement learning fine-tuning are conducted on a large language model to potentially achieve better results.
\end{enumerate}

% - Greedy is bad: not optimal
% - DP: intractible, plus requires perfect information about how the guesser would guess: impractical

% - Requires perfect information of how the guesser would guess

% Strategies that don't work: 
% - Blocker words
% - Zero-hint
% - Infinite-hint

% Add stochasticity to guessing

\section{Related works}
Prior works have explored both how to automate the spymaster and how to approximate how humans come up with the word associations used in the game Codenames.

\subsection{Pairwise Word Similarity}
A lot of the research in the existing literature focuses on improving the word similarity function used by both the automated spymasters and guessers. This is because an improvement to the word similarity function directly translates to an improvement in the performance of both spymasters and guessers.

All existing spymaster algorithms leverage a pairwise word similarity function $\sigma(w_i, w_j)$ or non-negative distance function $d(w_i, w_j)$ to suggest hints $h=(c, n)$ to the guessers, where $c$ denotes the clue word and $n$ denotes the number of words to guess. Prior work has explored $\sigma$ or $d$ functions derived from knowledge graphs \citep{kim-2019-codenames-greedy,koyyalagunta-2021-playing-codenames-language-graphs-word-embeddings}, the cosine similarity of text embeddings \citep{shen-etal-2018-comparing,kim-2019-codenames-greedy,Jaramillo_Charity_Canaan_Togelius_2020-codenames-transformers,koyyalagunta-2021-playing-codenames-language-graphs-word-embeddings,mills2022probing,cserhati-etal-2022-codenames-coocurrence,mills2022probing}, TF-IDF \citep{Jaramillo_Charity_Canaan_Togelius_2020-codenames-transformers}, and Naive Bayes \citep{Jaramillo_Charity_Canaan_Togelius_2020-codenames-transformers}. Similarity functions that directly leverage the frequency of the collocation of bigrams have also been used \citep{shen-etal-2018-comparing,mills2022probing}, as well as ones based off the related normalized pointwise mutual information (NMPI) metric \cite{cserhati-etal-2022-codenames-coocurrence}, which is closely related. Word embedding and co-occurrence metric approaches have been the most effective methods for approximating human word associations in Codenames. While word embedding approaches are more time- and space-efficient for calculating the similarity between different words, they occasionally fail to capture word associations from collocations \citep{smrz2019crowdsourcing}.

Some of the word embeddings that have been used for autonomous Codenames agents include Word2Vec \citep{mikolov2013efficient-word2vec}, GloVe \citep{pennington-etal-2014-glove}, fastText \citep{bojanowski-2016-fasttext}, Conceptnet Numberbatch \citep{speer2017conceptnet}, and Sentence-BERT \citep{reimers-2019-sentence-bert}, none of which are derived from human-labeled word associations. \cite{kumar-etal-2021-semantic-connector} constructed embeddings from several word association datasets, such as the Small World of Words (SWOW) dataset \citep{dedeyne-2018-swow} and the University of Southern Florida (USF) dataset \citep{nelson2004university-usf}, which each record the frequency of the first words that come to mind when people are presented with word prompts. \cite{kumar-etal-2021-semantic-connector} found that these embeddings produced word associations that were more similar to human chosen ones than word2vec and GloVe in a simplified version of Codenames.

% Most prior work on applying natural language processing to playing Codenames has 

% The earliest work on applying natural language processing to Codenames compared the use of bigram co-occurrence counts 

\subsection{Automating the Spymaster}
Let $M$ denote the set of words on the current player's team and $U=O\cup B\cup A$ denote the set of ``bad'' words, which includes words from the opposing team, bystander, and assassin words. Let $I_n$ denote the target or intended set of $n$ words on the board that is supposed to be targeted with the clue word $c$. When cosine similarity is used, the range of $\sigma$ is $[-1, 1]$. Cosine distance is calculated from the cosine similarity by $d(w_i, w_j) \coloneqq 1-\sigma(w_i, w_j)$.

In the literature, the spymasters choose words within the vocabulary $V$ which either maximize a scoring function $g(c, n)$ or minimize an energy function $f(c, n)$. \cite{kim-2019-codenames-greedy} uses the energy function
\begin{equation}
     f_{Kim}(c,n) = \begin{cases} 
      \max_{w\in I_n} d(c, w) & \text{if} \max_{w\in I_n} d(c, w) < \min_{u\in U} d(c, u) \\&\text{ and } \max_{w\in I_n} d(c, w) < \lambda_T\\
      \infty & \text{otherwise}
   \end{cases}
   \label{eq:f_kim}
\end{equation}
% \begin{equation}
%      f_{Kim}(c,n) = \begin{cases} 
%       \max_{w\in I_n} d(c, w) & \text{if} \max_{w\in I_n} d(c, w) < \min_{u\in U} d(c, u) \\&\text{ and } \max_{w\in I_n} d(c, w) < \lambda_T\\
%       \infty & \text{otherwise}
%    \end{cases}
%    \label{eq:g_kim}
% \end{equation}
where $\lambda_T$ is a parameter that governs how aggressive the spymaster is (\textit{low}=0.3, \textit{high}=0.7).
% In \cite{kim-2019-codenames-greedy}, $\max_{w\in I_n} d(c, w)$ is denoted as $d_r$ and $\min{u\in U} d(c, u)$ is denoted as $w_d$.

In all of the three final approaches mentioned in \cite{Jaramillo_Charity_Canaan_Togelius_2020-codenames-transformers}, they reweigh unintended words by -1 for bystanders, -2 for the opposing team's words, and -3 for the assassin, which I'll denote by $k(u)$ for $u \in U$. Let $\Vec{v}_{w}$ denote the embedding for word $w$ and let $csm(\Vec{v}, \Vec{x})=\frac{\Vec{v}\cdot \Vec{x}}{\norm{\Vec{v}}\norm{\Vec{x}}}$. Their best-performing algorithm leveraged word embeddings, with the scoring function 

\begin{equation}
    g_{Jara}(c, n) = csm\left(\Vec{c}, \frac{1}{\abs{I_n} + \abs{U}}(\sum_{w \in I_n}\Vec{w} + \sum_{u \in U}k(u)\Vec{u})\right)
    \label{eq:g_jara}
\end{equation}
% which is essentially the cosine similarity between the embedding of $c$ and the weighted centroid of the intended words $I_n$ and unintended words $U$. 
\cite{Jaramillo_Charity_Canaan_Togelius_2020-codenames-transformers} derived equation \eqref{eq:g_jara} by adding bad word reweighing to
\begin{equation}
    g_{mean} = csm(\Vec{c}, \frac{1}{\abs{I_n}}\sum_{w\in I_n}\Vec{w})
\end{equation}

\cite{koyyalagunta-2021-playing-codenames-language-graphs-word-embeddings} and \cite{cserhati-etal-2022-codenames-coocurrence} also define their own scoring functions, but only test them in a simplified Codenames environment with 20 words and just red and blue words.

% TODO: Change to we/our/us for final publication
Unlike prior works, I leave the bad word penalization to the RL agent and thus I do not directly penalize unintended words in the scoring function. Moreover, my RL environment is designed to work for a full game of Codenames. 

\subsection{Automating the Guessers}
All prior work automates the guessers by greedily choosing the word on the board that maximizes $\sigma(c, w)$ or that minimizes $d(c, w)$ for a given clue word $c$. Some works also looked at the correlation between the similarity of the scoring function and the human accuracy of guessing words \citep{kumar-etal-2021-semantic-connector} and \citep{cserhati-etal-2022-codenames-coocurrence}.

\section{Methodology}
% TODO: Change to we/our/us for final publication
To adapt a version Codenames for reinforcement learning, I first describe it as a Markov Decision Process (MDP), that is solely acted upon by the spymaster. I then describe how this MDP has been implemented as several OpenAI Gym \citep{brockman2016openai-gym} environments. I then describe the reinforcement learning (RL) agents that I will evaluate the environments on. Because convergence on a Codenames environment fails empirically, I will also describe a simplified Codenames environment, called ``ClickPixel''.
 % and the various experiments that for analyzing why the agents converge or not

\subsection{Formulation of Codenames as a MDP}
% Given a good similarity function $\sigma(c, w)$, the automated guessers can approximate human-chosen Codenames word associations quite well, using natural language processing. As a result, the guessers will be incorporated into the environment and not be directly controlled by the agents' actions.

Let $\mathcal B = M \cup U$, the set of 25 words on the board. Let $L$ denote the set of labels for the words on the board. Let $\ell\colon B\mapsto L$ be the oracle label function and let $m_t\colon B\mapsto \{0, 1\}$ be the masking function that indicates whether a word is chosen (1) or not (0) at step $t$. Let $\ell'_t$ be the normalized label function that relabels the ``blue'' and ``red'' labels to ``mine'' and ``opposing'' depending on whose turn it is at time step $t$. The state $s_t = (\mathcal B, m_t, \ell'_t)$ at time $t$.

In general, the action space for the spymasters is $A = V\times [0, 9]$, where $V$ is the vocabulary. Infinite word hints can be reduced to hints where $n$ is the number of words remaining, which is at most $9$. For the sake of simplicity, zero-word hints will not be handled for now, so $A = V\times [1, 9]$.

Between each step, all of the guessers for the team will make guesses based on the given action $a_t = (c_t, n_t)$ and until their turn ends. The guessers will guess the most similar $n_t$ words to the clue $c_t$ as long as $s(c, w) > \lambda$ for some hyperparameter $\lambda$, for each candidate word $w$, in descending order of similarity.

To match the scoring function used in \cite{kim-2019-codenames-greedy}, the agent will receive a reward of $-1$ for each turn when the agent has not won yet, $-25$ once the spymaster chooses the assassin word and $0$ once the spymaster wins. Thus, if the assassin word is never chosen, the final score is the negation of the number of turns it took for both teams to complete the board.

% \subsection{The reinforcement learning environment}
% Although prior techniques have used rule-based methods and supervised learning, none of them have used reinforcement learning to improve the agent.

% We will be using a modified version of the environment prop

% The agents will be evaluated on the mean 

\subsection{Codenames Gym Environment} \label{sec:gym}
% TODO: Change to we/our/us for final publication
I implement the above Codenames MDP as a a Gym \citep{brockman2016openai-gym} environment in my experiments as described below.
% Several methods for encoding the state space, action space, and scoring functions were implemented as configurable options when initializing the Codenames Gym environment. To ensure that the environment reflects a real environment with human guessers more closely, we introduce a novel guesser strategy in which stochasticity is incorporated into the order in which the guessers choose candidate words. Approximate nearest neighbour acceleration is also incorporated into the environment to accelerate how fast it runs so that the agents can finish training in reasonable period of time.

% The Gym environment is designed to be compatible with Stable Baselines 3 \citep{stable-baselines3}, which is a popular RL package with implementations for a few RL algorithms. 

\subsubsection{State Encoding Method}
% There are two ways that the words on the board are encoded: with (1) a pairwise word cosine similarity matrix and (2): with a one-hot encoding of the words on the board.

% TODO: Change to we/our/us for final publication
\paragraph{Pairwise Word Cosine Similarity Matrix (PWCSM)}
The state space is encoded as a $[-1, 1]^{j\abs{\mathcal B}+\abs{L}+2\times \abs{\mathcal B}}$ matrix, where $j$ is the number of word embeddings I want the agent to refer to. The first $j\abs{\mathcal B}$ rows encode $j$ similarity matrices encoding the pairwise cosine similarity of each word on the board. $j$ is set to 1 in practice, as no improvements were found when using different similarity matrices computed with different word embeddings.

\paragraph{One-Hot Word Encoding (OHWE)}
Assuming that $\abs{V} \equiv 0\;(\text{mod } \abs{\mathcal B})$, which is true for a default game of Codenames ($\abs{V} = 400$, $\abs{\mathcal B}=25$), the state space is encoded as a $[0, 1]^{\abs{V}/\abs{\mathcal B}+\abs{L}+2\times \abs{\mathcal B}}$ matrix. The first $\abs{V}/\abs{\mathcal B}$ rows are a wrapped one-hot membership vector for whether the words in the vocabulary are on the board.

% \subsubsection{Goal Encoding Method}
% To ensure that the environment supports hindsight experience replay (HER) \citep{andrychowicz2017hindsightexperiencereplay-her}, details about the intended goal are encoded into the environment. The goal is encoded as a $[0, 9]^{\abs{\mathcal{B}}\times 1}$ matrix where the first entry encodes the number of remaining words, the second entry encodes whether the other team has won yet and the third entry encodes whether the assassin word has been chosen. The rest of the entries in the vector are zeros. The reason why the vector must be encoded as a $\abs{\mathcal B} \times 1$ vector when only three dimensions are used is because it has to be possible to \texttt{vstack} the observation and the goal together.

\subsubsection{Action Encoding Method}
Let $G$ be the set of scoring functions used by the spymaster. The action is encoded as a $[0, 1]^{2\abs{\mathcal{B}} + \abs{G}\kappa \times 1}$ matrix, where $\kappa$ represents the number of word candidates generated. In other words, this is a one-hot vector of each word indicating which words were included in the target set $I_n$, which scoring strategy was used, and which of the $\kappa$ generated words were chosen by the spymaster.

\subsubsection{Spymaster Strategy}
The first scoring function used is $g_{mean}$, which was introduced by \cite{Jaramillo_Charity_Canaan_Togelius_2020-codenames-transformers}. The second is
\begin{equation}
     g_{minimax}(c,n) = \begin{cases} 
      \min_{w\in I_n} s(c, w) & \text{if} \min_{w\in I_n} s(c, w) > \lambda_T\\
      0 & \text{otherwise}
   \end{cases}
   \label{eq:f_kim}
\end{equation}
which is essentially $g_{Kim}$ except I remove the constraint on ensuring that bad words are not recommended, as I want the reinforcement learning algorithm to learn which actions are bad.
% TODO: Change to we/our/us for final publication (above)

\subsubsection{Guesser Strategy}
% TODO: Change to we/our/us for final publication
Although the environment can be configured to use (1) the greedy word selection strategy as in prior works, I also introduce (2) a novel stochastic guesser strategy. $n_t$ words are drawn without replacement from the distribution $softmax(\frac{s(c,\cdot)}{\tau})$, where the temperature $\tau$ is set to $0.05$ in practice. Empirically, the words are roughly chosen in the same order as the greedy approach, though words $w$ that have a close $s(c, w)$ value may have a slightly similar order.

% Thus, the environment has two guesser strategies: (1) the greedy strategy and (2) the stochastic guessing strategy described above.

\subsubsection{Word Embeddings}
Although the environment can support both GloVe and Conceptnet Numberbatch embeddings, the uncased GloVe-300d embeddings trained on 6 billion tokens from the Wikipedia 2014 and Gigaword 5 corpora are used for the experiments.  

\subsubsection{Approximate Nearest Neighbour Acceleration}
Because searching for the $n_t$ nearest neighbours to $I_n$ in $V$ according to the cosine similarity function is slow for large $V$ using exact distances, the Scalable Nearest Neighbour (ScaNN) algorithm \citep{guo-etal-2020-scann} is used to calculate the approximate $n_t$ nearest neighbours according to the cosine similarity. Using ScaNN to accelerate the environment speeds up the training from about 1-2 seconds per iteration to about 65-80 iterations per second.

\subsection{RL Agents}
% TODO: Use we/our/us for final publication
I use the soft actor critic (SAC) \citep{haarnoja-2018-sac}, SAC with hindsight experience replay (HER) \citep{andrychowicz2017hindsightexperiencereplay-her}, proximal policy optimization (PPO) \citep{schulman-etal-2017-ppo}, and advantage actor critic (A2C) \citep{mnih-etal-2016-a2c} agents implemented in Stable Baselines 3 \citep{stable-baselines3} for my experiments.

\subsection{ClickPixel Gym Environment}
To figure out why the agents failed to converge on the Codenames environment, a simplified environment called ``ClickPixel'' was created.
In ClickPixel, a single pixel is highlighted in a board of $q\times q$ pixels. The rest of the pixels are unhighlighted by default and are highlighted when chosen by the agent. Because only PPO and A2C support discrete action spaces in Stable Baselines 3, only these two agents are evaluated on ClickPixel.

\subsection{Whack-a-Mole Gym Environment}
Since in ClickPixel, the agent is required to select the a single grid on the board, whereas in Codenames, the agent is required to select a subset of the grids on the board, we also introduce the Whack-a-Mole environment, where the agent has to select the highlighted subset of words at each time step. The agent is rewarded with the mean classification accuracy of choosing or not choosing each word shown on the board.

% \subsection{Single-player empirical simulation}
% - Must be in C++ to be fast
% - Open Spiel

% Speedups with ScANN and caching
% <- indicate the speedups: calculate it!

% Various levels of recall

% Also: coordconv problem:
% -> surprised it takes so long to learn the identity function

% Extended coordconv:
% -> once you get feedback, it fails miserably

% \subsection{Multi-player empirical simulation}
% \subsubsection{Self-play}
% Blah
% \subsubsection{Versus other bots}
% Blah
% \subsubsection{Human evaluation}
% Blah

\section{Results}
% Compare empirically the techniques for complexity, performance, ease of use, etc.
% TODO: Change to we/our/us for final publication
I find that none of the Codenames environments converged after being trained for 100k steps. In fact, a random policy sampled from the action space had produced a return ($-32.00 \pm 3.97$) that was similar to the return produced by all four agents and a greedy strategy choosing one word at a time, using the first word recommended from the $g_{mean}$ scoring function had a mean return of $-10.4\pm 5.9$. However, when using a $2\times 2$ board with a word of each colour, all four agents converge when using a PWCSM environment and both PPO and A2C converge on a OHWE environment, though neither of the SAC agents converge.

% \begin{figure}[htbp]
% \centering
%   \includegraphics[width=.9\linewidth]{assets/images/codenames_env_old_plot.png}
%   \caption{None of the four agents converge.}
%   \label{fig:training_plot}
% \end{figure}

\subsection{ClickPixel}
% To figure out why the agents failed to converge on the Codenames environment, a simplified environment called ``ClickPixel'' was created.
% In ClickPixel, a single pixel is highlighted in a board of $q\times q$ pixels. The rest of the pixels are unhighlighted by default and are highlighted when chosen by the agent. Because only PPO and A2C support discrete action spaces in Stable Baselines 3, only these two agents are evaluated on ClickPixel.

Interestingly enough, while both PPO and A2C converge when $q=2$, only A2C converges when $q=5$ (the width of a board in Codenames) and neither agent converges when $q=36$ (the minimum dimensions required for CNNs to be used as the encoding head in Stable Baselines 3), which highlights the difficulty of RL algorithms converging to an optimal policy from a randomly initialized one when the action space is large.

Surprisingly, frame stacking (4 consecutive frames) hurts convergence and giving the RL agents only one move to make the correct action hurts how quickly the agents learn the optimal policy, which is surprising given how ClickPixel can be reduced to learning the identity function with a neural network.

\subsection{Whack-a-Mole}
While A2C converges in Whack-a-Mole when $q=2$, it fails to converge for $q=5$ and above. Moreover, PPO fails to converge in both cases, with a mean classification return of $48.45\pm0.87$, over 99 steps, which is close to a 50\% accuracy from guessing randomly in binary classification.

% \section{Discussion}

\section{Limitations}
In this paper, I formulate Codenames as a single-agent environment, but a true game of Codenames is a zero-sum game with two players. Thus, an agent that learns an optimal policy in my environment might give a hint that makes it easy for the opponent to suggest a hint that takes a lot of their words off the board, whereas in a true game of Codenames, the spymasters would not be incentivized to do so.

\section{Future Work}
As for future research, I would recommend trying to collect a large dataset of human Codenames games and learning a scoring function using this database of Codenames games. Then, by combining self-play while regularizing the learned policy to be similar to a learned human policy using KL-Divergence, a strong human-like policy can be obtained \citep{pmlr-v162-jacob22a-pikl,bakhtin-etal-2022-cicero,bakhtin-etal-2022-dil-pikl}.

Another approach would be to fine-tune a language model such as FLAN-T5 \citep{chung2022scaling-flan-t5} on predicting properties about synthetic Codenames games that are encoded using a text template. Reinforcement learning from human feedback (RHLF) \citep{christiano2017deep-rhlf} can then be used to further fine-tune the LLM to play as an effective Codenames spymaster.

% Moreover, a better similarity function can be designed that takes the entire target set and the words on the board into consideration, such as $s(c, I_n, \mathcal B)$.

\section{Conclusion}
% Conclusion:
% What is the best technique?
% Is any technique good enough to declare the problem solved?
% What future research do you recommend?

So far, the best technique is A2C, as evaluated on the ClickPixel environment. None of the techniques evaluated are good enough to declare the problem solved, as none of SAC, PPO, and A2C were able to converge to a policy that beat a random policy or a naive greedy policy evaluated on the environment. 

% Acknowledgements should go at the end, before appendices and references

\acks{I would like to thank Pascal Poupart for providing feedback for my research project.}
% \acks{We would like to acknowledge support for this project
% from the National Science Foundation (NSF grant IIS-9988642)
% and the Multidisciplinary Research Program of the Department
% of Defense (MURI N00014-00-1-0637). }

% % Manual newpage inserted to improve layout of sample file - not
% % needed in general before appendices/bibliography.

% \newpage

\appendix
\section{Speculative Discussion}
Because what each action does is completely dependent on what words are on the board, I hypothesize that this causes the agents to have a harder time finding an optimal policy.

\section{Goal Encoding Method}
\label{app:goal_encoding}
To ensure that the Codenames environment supports hindsight experience replay (HER) \citep{andrychowicz2017hindsightexperiencereplay-her}, details about the intended goal are encoded into the environment. The goal is encoded as a $[0, 9]^{\abs{\mathcal{B}}\times 1}$ matrix where the first entry encodes the number of remaining words, the second entry encodes whether the other team has won yet and the third entry encodes whether the assassin word has been chosen. The rest of the entries in the vector are zeros. The reason why the vector must be encoded as a $\abs{\mathcal B} \times 1$ vector when only three dimensions are used is because it has to be possible to \texttt{vstack} the observation and the goal together.

\section{Experiment Plots}
\label{app:exp_plots}
Some of the experiment plots for the Codenames, ClickPixel, and Whack-a-Mole environments are included below. 

Because the plots for each Codenames experiment are almost completely identical, only the plot for a PWCSM experiment is included below.

\subsection{Codenames}
\begin{figure}[htbp]
\centering
  \includegraphics[width=.9\linewidth]{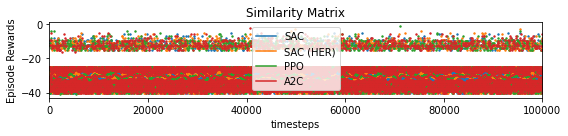}
  \caption{The episode rewards from training all four agents with 100k steps.}
  \label{fig:codenames-csm-plot}
\end{figure}

None of the four agents converge when trained on a Codenames environment for a $5\times 5$ board, as shown in Figure~\ref{fig:codenames-csm-plot}.

\subsection{ClickPixel}
Unfortunately, the models had a difficult time learning an optimal policy for ClickPixel, even when frame stacking and reward shaping were added.
% ClickPixel was also known as Gridtest in earlier debugging plots.

\begin{figure}[htbp]
\centering
  \includegraphics[width=.9\linewidth]{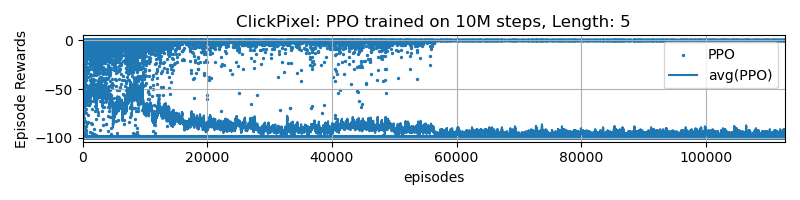}
  \includegraphics[width=.9\linewidth]{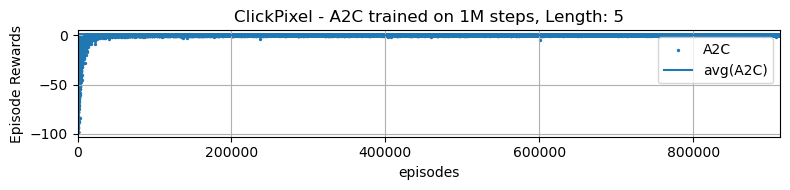}
  \caption{Training PPO on the ClickPixel environment for 10M steps doesn't cause it to converge to a good policy. In fact, it learns the worst policy. Neither PPO nor A2C learn a better policy when using a $5\times 5$ board.}
  \label{fig:gridtest-l5}
\end{figure}

\begin{figure}[htbp]
\centering
  \includegraphics[width=.9\linewidth]{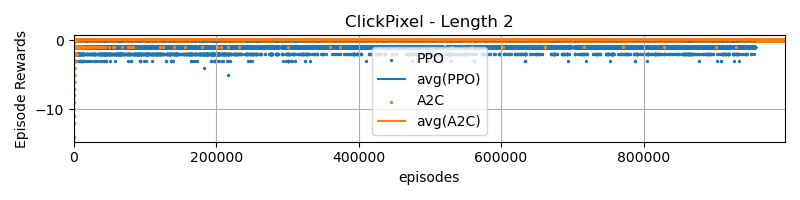}
  \caption{However, both agents can learn an optimal policy for a smaller $2\times 2$ board.}
  \label{fig:gridtest-l2}
\end{figure}

\begin{figure}[htbp]
\centering
  \includegraphics[width=.9\linewidth]{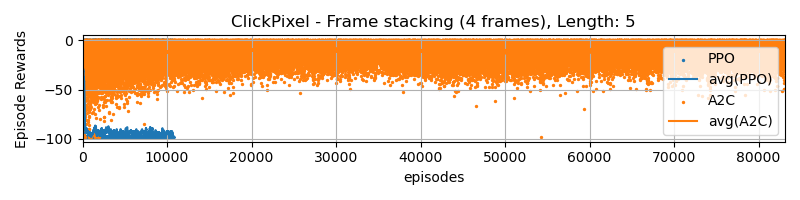}
  \caption{Frame stacking actually hurts the performance of the algorithms, surprisingly.}
  \label{fig:gridtest-frame-stack-l5}
\end{figure}

\begin{figure}[htbp]
\centering
  \includegraphics[width=.9\linewidth]{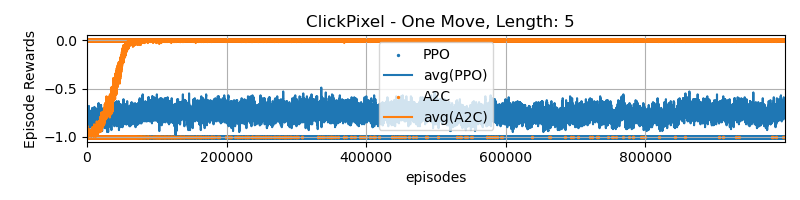}
  \caption{Giving the agent one move in an episode to make the right decision causes A2C to take longer to converge and doesn't help PPO at all.}
  \label{fig:gridtest-onemove-l5}
\end{figure}

\begin{figure}[htbp]
\centering
  \includegraphics[width=.9\linewidth]{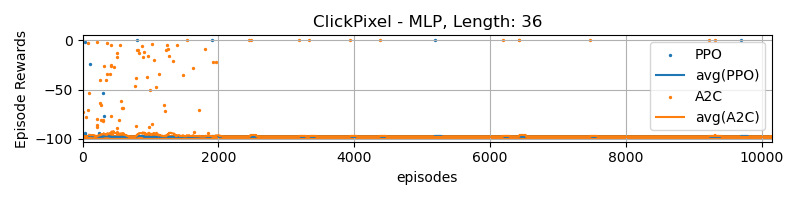}
    \includegraphics[width=.9\linewidth]{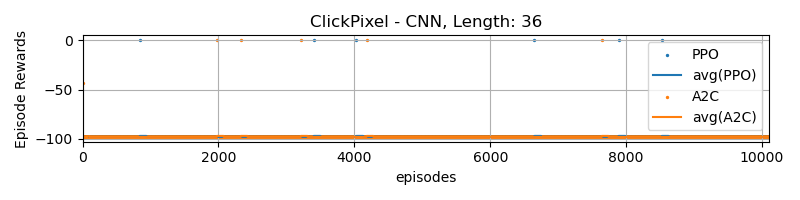}
  \caption{As expected, increasing the board size to $36\times 36$ doesn't help the agents at all. Less expected is the fact that switching to a CNN policy head doesn't  help the agents learn a good policy.}
  \label{fig:gridtest-l36}
\end{figure}

\begin{figure}[htbp]
\centering
\includegraphics[width=.9\linewidth]{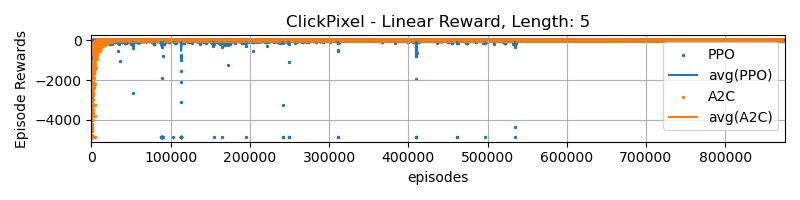}
\includegraphics[width=.9\linewidth]{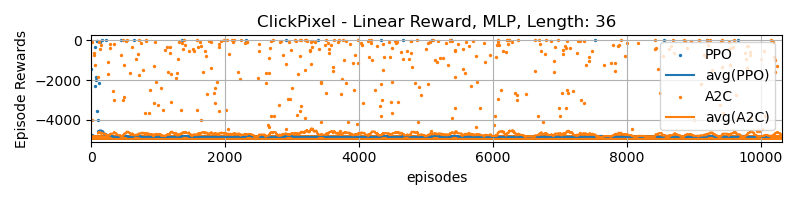}
\includegraphics[width=.9\linewidth]{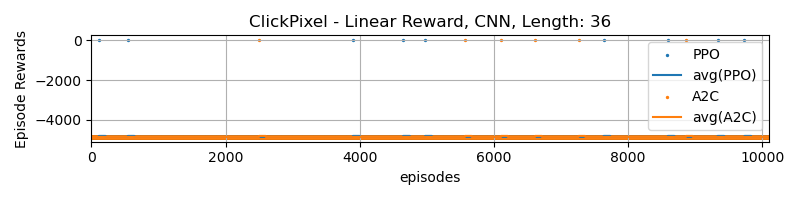}
  \caption{Although making the negative per-step reward proportional to the number of steps taken helps the PPO agent learn a better policy on a $5\times 5$ board, it doesn't scale up to get the agents to converge to a good policy on a $36\times 36$ board.}
  \label{fig:gridtest-linear-reward}
\end{figure}

\subsection{Whack-a-Mole}
\begin{figure}[htbp]
\centering
  \includegraphics[width=.9\linewidth]{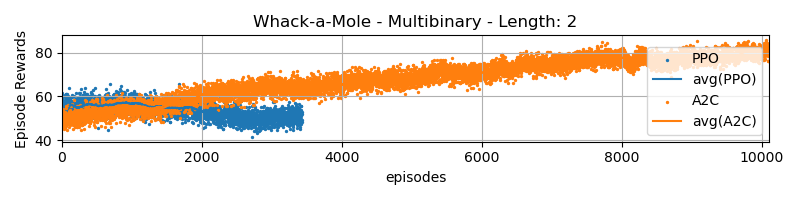}
  \caption{A2C is able to learn a better policy when using a $2\times 2$ board.}
  \label{fig:whack-l2}
\end{figure}

\begin{figure}[htbp]
\centering
  \includegraphics[width=.9\linewidth]{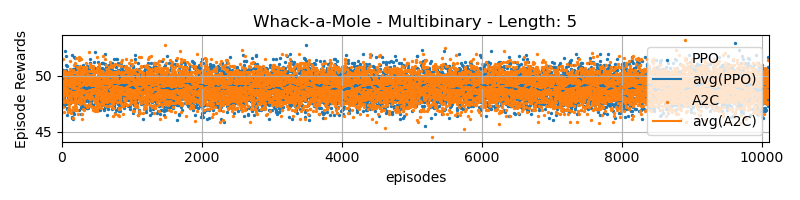}
  \caption{Neither PPO nor A2C learn a better policy when using a $5\times 5$ board.}
  \label{fig:whack-l5}
\end{figure}

A2C was able to learn a good policy on a $2\times 2$ board as shown in Figure~\ref{fig:whack-l2}, but not on the larger $5\times 5$ board, as shown in Figure~\ref{fig:whack-l5}.

% \newpage

% % Note: in this sample, the section number is hard-coded in. Following
% % proper LaTeX conventions, it should properly be coded as a reference:

% %In this appendix we prove the following theorem from
% %Section~\ref{sec:textree-generalization}:

% In this appendix we prove the following theorem from
% Section~6.2:

% \noindent
% {\bf Theorem} {\it Let $u,v,w$ be discrete variables such that $v, w$ do
% not co-occur with $u$ (i.e., $u\neq0\;\Rightarrow \;v=w=0$ in a given
% dataset $\dataset$). Let $N_{v0},N_{w0}$ be the number of data points for
% which $v=0, w=0$ respectively, and let $I_{uv},I_{uw}$ be the
% respective empirical mutual information values based on the sample
% $\dataset$. Then
% \[
% 	N_{v0} \;>\; N_{w0}\;\;\Rightarrow\;\;I_{uv} \;\leq\;I_{uw}
% \]
% with equality only if $u$ is identically 0.} \hfill\BlackBox

% \noindent
% {\bf Proof}. We use the notation:
% \[
% P_v(i) \;=\;\frac{N_v^i}{N},\;\;\;i \neq 0;\;\;\;
% P_{v0}\;\equiv\;P_v(0)\; = \;1 - \sum_{i\neq 0}P_v(i).
% \]
% These values represent the (empirical) probabilities of $v$
% taking value $i\neq 0$ and 0 respectively.  Entropies will be denoted
% by $H$. We aim to show that $\fracpartial{I_{uv}}{P_{v0}} < 0$....\\

% {\noindent \em Remainder omitted in this sample. See http://www.jmlr.org/papers/ for full paper.}

\vskip 0.2in
\bibliography{sample}

\begin{thebibliography}{38}
\providecommand{\natexlab}[1]{#1}
\providecommand{\url}[1]{\texttt{#1}}
\expandafter\ifx\csname urlstyle\endcsname\relax
  \providecommand{\doi}[1]{doi: #1}\else
  \providecommand{\doi}{doi: \begingroup \urlstyle{rm}\Url}\fi

\bibitem[Andrychowicz et~al.(2017)Andrychowicz, Wolski, Ray, Schneider, Fong,
  Welinder, McGrew, Tobin, Pieter~Abbeel, and
  Zaremba]{andrychowicz2017hindsightexperiencereplay-her}
Marcin Andrychowicz, Filip Wolski, Alex Ray, Jonas Schneider, Rachel Fong,
  Peter Welinder, Bob McGrew, Josh Tobin, OpenAI Pieter~Abbeel, and Wojciech
  Zaremba.
\newblock Hindsight experience replay.
\newblock \emph{Advances in neural information processing systems}, 30, 2017.

\bibitem[Bakhtin et~al.(2021)Bakhtin, Wu, Lerer, and
  Brown]{bakhtin-2021-no-press-diplomacy-fail}
Anton Bakhtin, David Wu, Adam Lerer, and Noam Brown.
\newblock No-press diplomacy from scratch, 2021.
\newblock URL \url{https://arxiv.org/abs/2110.02924}.

\bibitem[Bakhtin et~al.(2022{\natexlab{a}})Bakhtin, Brown, Dinan, Farina,
  Flaherty, Fried, Goff, Gray, Hu, Jacob, Komeili, Konath, Kwon, Lerer, Lewis,
  Miller, Mitts, Renduchintala, Roller, Rowe, Shi, Spisak, Wei, Wu, Zhang, and
  Zijlstra]{bakhtin-etal-2022-cicero}
Anton Bakhtin, Noam Brown, Emily Dinan, Gabriele Farina, Colin Flaherty, Daniel
  Fried, Andrew Goff, Jonathan Gray, Hengyuan Hu, Athul~Paul Jacob, Mojtaba
  Komeili, Karthik Konath, Minae Kwon, Adam Lerer, Mike Lewis, Alexander~H.
  Miller, Sasha Mitts, Adithya Renduchintala, Stephen Roller, Dirk Rowe, Weiyan
  Shi, Joe Spisak, Alexander Wei, David Wu, Hugh Zhang, and Markus Zijlstra.
\newblock Human-level play in the game of <i>diplomacy</i> by combining
  language models with strategic reasoning.
\newblock \emph{Science}, 378\penalty0 (6624):\penalty0 1067--1074,
  2022{\natexlab{a}}.
\newblock \doi{10.1126/science.ade9097}.
\newblock URL \url{https://www.science.org/doi/abs/10.1126/science.ade9097}.

\bibitem[Bakhtin et~al.(2022{\natexlab{b}})Bakhtin, Wu, Lerer, Gray, Jacob,
  Farina, Miller, and Brown]{bakhtin-etal-2022-dil-pikl}
Anton Bakhtin, David~J Wu, Adam Lerer, Jonathan Gray, Athul~Paul Jacob,
  Gabriele Farina, Alexander~H Miller, and Noam Brown.
\newblock Mastering the game of no-press diplomacy via human-regularized
  reinforcement learning and planning, 2022{\natexlab{b}}.
\newblock URL \url{https://arxiv.org/abs/2210.05492}.

\bibitem[Bojanowski et~al.(2016)Bojanowski, Grave, Joulin, and
  Mikolov]{bojanowski-2016-fasttext}
Piotr Bojanowski, Edouard Grave, Armand Joulin, and Tomas Mikolov.
\newblock Enriching word vectors with subword information, 2016.
\newblock URL \url{https://arxiv.org/abs/1607.04606}.

\bibitem[Brockman et~al.(2016)Brockman, Cheung, Pettersson, Schneider,
  Schulman, Tang, and Zaremba]{brockman2016openai-gym}
Greg Brockman, Vicki Cheung, Ludwig Pettersson, Jonas Schneider, John Schulman,
  Jie Tang, and Wojciech Zaremba.
\newblock Openai gym.
\newblock \emph{arXiv preprint arXiv:1606.01540}, 2016.

\bibitem[Christiano et~al.(2017)Christiano, Leike, Brown, Martic, Legg, and
  Amodei]{christiano2017deep-rhlf}
Paul~F Christiano, Jan Leike, Tom Brown, Miljan Martic, Shane Legg, and Dario
  Amodei.
\newblock Deep reinforcement learning from human preferences.
\newblock \emph{Advances in neural information processing systems}, 30, 2017.

\bibitem[Chung et~al.(2022)Chung, Hou, Longpre, Zoph, Tay, Fedus, Li, Wang,
  Dehghani, Brahma, et~al.]{chung2022scaling-flan-t5}
Hyung~Won Chung, Le~Hou, Shayne Longpre, Barret Zoph, Yi~Tay, William Fedus,
  Eric Li, Xuezhi Wang, Mostafa Dehghani, Siddhartha Brahma, et~al.
\newblock Scaling instruction-finetuned language models.
\newblock \emph{arXiv preprint arXiv:2210.11416}, 2022.

\bibitem[Chvátil(2015)]{codenamesrules}
Vlaada Chvátil.
\newblock Codenames rules - czech games edition, 2015.
\newblock URL \url{https://czechgames.com/files/rules/codenames-rules-en.pdf}.

\bibitem[C{\^o}t{\'e} et~al.(2018)C{\^o}t{\'e}, K{\'a}d{\'a}r, Yuan, Kybartas,
  Barnes, Fine, Moore, Hausknecht, El~Asri, Adada, et~al.]{cote2018textworld}
Marc-Alexandre C{\^o}t{\'e}, Akos K{\'a}d{\'a}r, Xingdi Yuan, Ben Kybartas,
  Tavian Barnes, Emery Fine, James Moore, Matthew Hausknecht, Layla El~Asri,
  Mahmoud Adada, et~al.
\newblock Textworld: A learning environment for text-based games.
\newblock In \emph{Workshop on Computer Games}, pages 41--75. Springer, 2018.

\bibitem[Cserh{\'a}ti et~al.(2022)Cserh{\'a}ti, Kollath, Kicsi, and
  Berend]{cserhati-etal-2022-codenames-coocurrence}
R{\'e}ka Cserh{\'a}ti, Istvan Kollath, Andr{\'a}s Kicsi, and G{\'a}bor Berend.
\newblock Codenames as a game of co-occurrence counting.
\newblock In \emph{Proceedings of the Workshop on Cognitive Modeling and
  Computational Linguistics}, pages 43--53, Dublin, Ireland, May 2022.
  Association for Computational Linguistics.
\newblock \doi{10.18653/v1/2022.cmcl-1.5}.
\newblock URL \url{https://aclanthology.org/2022.cmcl-1.5}.

\bibitem[De~Deyne et~al.(2018)De~Deyne, Navarro, Perfors, Brysbaert, and
  Storms]{dedeyne-2018-swow}
Simon De~Deyne, Danielle Navarro, Amy Perfors, Marc Brysbaert, and Gert Storms.
\newblock The “small world of words” english word association norms for
  over 12,000 cue words.
\newblock \emph{Behavior Research Methods}, 51, 10 2018.
\newblock \doi{10.3758/s13428-018-1115-7}.

\bibitem[Guo et~al.(2020)Guo, Sun, Lindgren, Geng, Simcha, Chern, and
  Kumar]{guo-etal-2020-scann}
Ruiqi Guo, Philip Sun, Erik Lindgren, Quan Geng, David Simcha, Felix Chern, and
  Sanjiv Kumar.
\newblock Accelerating large-scale inference with anisotropic vector
  quantization.
\newblock In Hal~Daumé III and Aarti Singh, editors, \emph{Proceedings of the
  37th International Conference on Machine Learning}, volume 119 of
  \emph{Proceedings of Machine Learning Research}, pages 3887--3896. PMLR,
  13--18 Jul 2020.
\newblock URL \url{https://proceedings.mlr.press/v119/guo20h.html}.

\bibitem[Haarnoja et~al.(2018)Haarnoja, Zhou, Abbeel, and
  Levine]{haarnoja-2018-sac}
Tuomas Haarnoja, Aurick Zhou, Pieter Abbeel, and Sergey Levine.
\newblock Soft actor-critic: Off-policy maximum entropy deep reinforcement
  learning with a stochastic actor, 2018.
\newblock URL \url{https://arxiv.org/abs/1801.01290}.

\bibitem[Jacob et~al.(2022)Jacob, Wu, Farina, Lerer, Hu, Bakhtin, Andreas, and
  Brown]{pmlr-v162-jacob22a-pikl}
Athul~Paul Jacob, David~J Wu, Gabriele Farina, Adam Lerer, Hengyuan Hu, Anton
  Bakhtin, Jacob Andreas, and Noam Brown.
\newblock Modeling strong and human-like gameplay with {KL}-regularized search.
\newblock In Kamalika Chaudhuri, Stefanie Jegelka, Le~Song, Csaba Szepesvari,
  Gang Niu, and Sivan Sabato, editors, \emph{Proceedings of the 39th
  International Conference on Machine Learning}, volume 162 of
  \emph{Proceedings of Machine Learning Research}, pages 9695--9728. PMLR,
  17--23 Jul 2022.
\newblock URL \url{https://proceedings.mlr.press/v162/jacob22a.html}.

\bibitem[Jaderberg et~al.(2019)Jaderberg, Czarnecki, Dunning, Marris, Lever,
  Castañeda, Beattie, Rabinowitz, Morcos, Ruderman, Sonnerat, Green, Deason,
  Leibo, Silver, Hassabis, Kavukcuoglu, and Graepel]{jaderberg2019human}
Max Jaderberg, Wojciech~M. Czarnecki, Iain Dunning, Luke Marris, Guy Lever,
  Antonio~Garcia Castañeda, Charles Beattie, Neil~C. Rabinowitz, Ari~S.
  Morcos, Avraham Ruderman, Nicolas Sonnerat, Tim Green, Louise Deason, Joel~Z.
  Leibo, David Silver, Demis Hassabis, Koray Kavukcuoglu, and Thore Graepel.
\newblock Human-level performance in 3d multiplayer games with population-based
  reinforcement learning.
\newblock \emph{Science}, 364\penalty0 (6443):\penalty0 859--865, 2019.
\newblock \doi{10.1126/science.aau6249}.
\newblock URL \url{https://www.science.org/doi/abs/10.1126/science.aau6249}.

\bibitem[Jaramillo et~al.(2020)Jaramillo, Charity, Canaan, and
  Togelius]{Jaramillo_Charity_Canaan_Togelius_2020-codenames-transformers}
Catalina Jaramillo, Megan Charity, Rodrigo Canaan, and Julian Togelius.
\newblock Word autobots: Using transformers for word association in the game
  codenames.
\newblock \emph{Proceedings of the AAAI Conference on Artificial Intelligence
  and Interactive Digital Entertainment}, 16\penalty0 (1):\penalty0 231--237,
  Oct. 2020.
\newblock \doi{10.1609/aiide.v16i1.7435}.
\newblock URL \url{https://ojs.aaai.org/index.php/AIIDE/article/view/7435}.

\bibitem[Kim et~al.(2019)Kim, Ruzmaykin, Truong, and
  Summerville]{kim-2019-codenames-greedy}
Andrew Kim, Maxim Ruzmaykin, Aaron Truong, and Adam Summerville.
\newblock Cooperation and codenames: Understanding natural language processing
  via codenames.
\newblock \emph{Proceedings of the AAAI Conference on Artificial Intelligence
  and Interactive Digital Entertainment}, 15\penalty0 (1):\penalty0 160--166,
  Oct. 2019.
\newblock \doi{10.1609/aiide.v15i1.5239}.
\newblock URL \url{https://ojs.aaai.org/index.php/AIIDE/article/view/5239}.

\bibitem[Koyyalagunta et~al.(2021)Koyyalagunta, Sun, Draelos, and
  Rudin]{koyyalagunta-2021-playing-codenames-language-graphs-word-embeddings}
Divya Koyyalagunta, Anna Sun, Rachel~Lea Draelos, and Cynthia Rudin.
\newblock Playing codenames with language graphs and word embeddings.
\newblock \emph{J. Artif. Intell. Res.}, 71:\penalty0 319--346, 2021.

\bibitem[Kumar et~al.(2021)Kumar, Steyvers, and
  Balota]{kumar-etal-2021-semantic-connector}
Abhilasha~A. Kumar, Mark Steyvers, and David~A. Balota.
\newblock Semantic memory search and retrieval in a novel cooperative word
  game: A comparison of associative and distributional semantic models.
\newblock \emph{Cognitive Science}, 45\penalty0 (10):\penalty0 e13053, 2021.
\newblock \doi{https://doi.org/10.1111/cogs.13053}.
\newblock URL \url{https://onlinelibrary.wiley.com/doi/abs/10.1111/cogs.13053}.

\bibitem[Lerer et~al.(2020)Lerer, Hu, Foerster, and Brown]{lerer2020improving}
Adam Lerer, Hengyuan Hu, Jakob Foerster, and Noam Brown.
\newblock Improving policies via search in cooperative partially observable
  games.
\newblock In \emph{Proceedings of the AAAI Conference on Artificial
  Intelligence}, volume~34, pages 7187--7194, 2020.

\bibitem[Mikolov et~al.(2013)Mikolov, Chen, Corrado, and
  Dean]{mikolov2013efficient-word2vec}
Tomas Mikolov, Kai Chen, Greg Corrado, and Jeffrey Dean.
\newblock Efficient estimation of word representations in vector space.
\newblock \emph{arXiv preprint arXiv:1301.3781}, 2013.

\bibitem[Mills(2022)]{mills2022probing}
Tracey Mills.
\newblock Probing nlp conceptual relatedness judgments through the word-based
  board game codenames.
\newblock 2022.

\bibitem[Mnih et~al.(2013)Mnih, Kavukcuoglu, Silver, Graves, Antonoglou,
  Wierstra, and Riedmiller]{mnih2013playing}
Volodymyr Mnih, Koray Kavukcuoglu, David Silver, Alex Graves, Ioannis
  Antonoglou, Daan Wierstra, and Martin Riedmiller.
\newblock Playing atari with deep reinforcement learning.
\newblock \emph{arXiv preprint arXiv:1312.5602}, 2013.

\bibitem[Mnih et~al.(2016)Mnih, Badia, Mirza, Graves, Lillicrap, Harley,
  Silver, and Kavukcuoglu]{mnih-etal-2016-a2c}
Volodymyr Mnih, Adrià~Puigdomènech Badia, Mehdi Mirza, Alex Graves,
  Timothy~P. Lillicrap, Tim Harley, David Silver, and Koray Kavukcuoglu.
\newblock Asynchronous methods for deep reinforcement learning.
\newblock 2016.
\newblock \doi{10.48550/ARXIV.1602.01783}.
\newblock URL \url{https://arxiv.org/abs/1602.01783}.

\bibitem[Nelson et~al.(2004)Nelson, McEvoy, and
  Schreiber]{nelson2004university-usf}
Douglas~L Nelson, Cathy~L McEvoy, and Thomas~A Schreiber.
\newblock The university of south florida free association, rhyme, and word
  fragment norms.
\newblock \emph{Behavior Research Methods, Instruments, \& Computers},
  36\penalty0 (3):\penalty0 402--407, 2004.

\bibitem[Ouyang et~al.(2022)Ouyang, Wu, Jiang, Almeida, Wainwright, Mishkin,
  Zhang, Agarwal, Slama, Ray, et~al.]{ouyang2022training-instructgpt}
Long Ouyang, Jeff Wu, Xu~Jiang, Diogo Almeida, Carroll~L Wainwright, Pamela
  Mishkin, Chong Zhang, Sandhini Agarwal, Katarina Slama, Alex Ray, et~al.
\newblock Training language models to follow instructions with human feedback.
\newblock \emph{arXiv preprint arXiv:2203.02155}, 2022.

\bibitem[Pennington et~al.(2014)Pennington, Socher, and
  Manning]{pennington-etal-2014-glove}
Jeffrey Pennington, Richard Socher, and Christopher Manning.
\newblock {G}lo{V}e: Global vectors for word representation.
\newblock In \emph{Proceedings of the 2014 Conference on Empirical Methods in
  Natural Language Processing ({EMNLP})}, pages 1532--1543, Doha, Qatar,
  October 2014. Association for Computational Linguistics.
\newblock \doi{10.3115/v1/D14-1162}.
\newblock URL \url{https://aclanthology.org/D14-1162}.

\bibitem[Raffin et~al.(2021)Raffin, Hill, Gleave, Kanervisto, Ernestus, and
  Dormann]{stable-baselines3}
Antonin Raffin, Ashley Hill, Adam Gleave, Anssi Kanervisto, Maximilian
  Ernestus, and Noah Dormann.
\newblock Stable-baselines3: Reliable reinforcement learning implementations.
\newblock \emph{Journal of Machine Learning Research}, 22\penalty0
  (268):\penalty0 1--8, 2021.
\newblock URL \url{http://jmlr.org/papers/v22/20-1364.html}.

\bibitem[Reimers and Gurevych(2019)]{reimers-2019-sentence-bert}
Nils Reimers and Iryna Gurevych.
\newblock Sentence-bert: Sentence embeddings using siamese bert-networks.
\newblock In \emph{Proceedings of the 2019 Conference on Empirical Methods in
  Natural Language Processing}. Association for Computational Linguistics, 11
  2019.
\newblock URL \url{https://arxiv.org/abs/1908.10084}.

\bibitem[Schulman et~al.(2017)Schulman, Wolski, Dhariwal, Radford, and
  Klimov]{schulman-etal-2017-ppo}
John Schulman, Filip Wolski, Prafulla Dhariwal, Alec Radford, and Oleg Klimov.
\newblock Proximal policy optimization algorithms, 2017.
\newblock URL \url{https://arxiv.org/abs/1707.06347}.

\bibitem[Shen et~al.(2018)Shen, Hofer, Felbo, and
  Levy]{shen-etal-2018-comparing}
Judy~Hanwen Shen, Matthias Hofer, Bjarke Felbo, and Roger Levy.
\newblock Comparing models of associative meaning: An empirical investigation
  of reference in simple language games.
\newblock In \emph{Proceedings of the 22nd Conference on Computational Natural
  Language Learning}, pages 292--301, Brussels, Belgium, October 2018.
  Association for Computational Linguistics.
\newblock \doi{10.18653/v1/K18-1029}.
\newblock URL \url{https://aclanthology.org/K18-1029}.

\bibitem[Silver et~al.(2016)Silver, Huang, Maddison, Guez, Sifre, van~den
  Driessche, Schrittwieser, Antonoglou, Panneershelvam, Lanctot, Dieleman,
  Grewe, Nham, Kalchbrenner, Sutskever, Lillicrap, Leach, Kavukcuoglu, Graepel,
  and Hassabis]{silver2016alphago}
David Silver, Aja Huang, Christopher~J. Maddison, Arthur Guez, Laurent Sifre,
  George van~den Driessche, Julian Schrittwieser, Ioannis Antonoglou, Veda
  Panneershelvam, Marc Lanctot, Sander Dieleman, Dominik Grewe, John Nham, Nal
  Kalchbrenner, Ilya Sutskever, Timothy Lillicrap, Madeleine Leach, Koray
  Kavukcuoglu, Thore Graepel, and Demis Hassabis.
\newblock Mastering the game of go with deep neural networks and tree search.
\newblock \emph{Nature}, 529:\penalty0 484--503, 2016.
\newblock URL
  \url{http://www.nature.com/nature/journal/v529/n7587/full/nature16961.html}.

\bibitem[Silver et~al.(2018)Silver, Hubert, Schrittwieser, Antonoglou, Lai,
  Guez, Lanctot, Sifre, Kumaran, Graepel, et~al.]{silver2018general-alphazero}
David Silver, Thomas Hubert, Julian Schrittwieser, Ioannis Antonoglou, Matthew
  Lai, Arthur Guez, Marc Lanctot, Laurent Sifre, Dharshan Kumaran, Thore
  Graepel, et~al.
\newblock A general reinforcement learning algorithm that masters chess, shogi,
  and go through self-play.
\newblock \emph{Science}, 362\penalty0 (6419):\penalty0 1140--1144, 2018.

\bibitem[Siu et~al.(2021)Siu, Pena, Chen, Zhou, Lopez, Palko, Chang, and
  Allen]{siu-2021-hanabi-fail}
Ho~Chit Siu, Jaime~D. Pena, Edenna Chen, Yutai Zhou, Victor~J. Lopez, Kyle
  Palko, Kimberlee~C. Chang, and Ross~E. Allen.
\newblock Evaluation of human-ai teams for learned and rule-based agents in
  hanabi, 2021.
\newblock URL \url{https://arxiv.org/abs/2107.07630}.

\bibitem[Smrz(2019)]{smrz2019crowdsourcing}
Pavel Smrz.
\newblock Crowdsourcing complex associations among words by means of a game.
\newblock In \emph{CS \& IT Conference Proceedings}, volume~9. CS \& IT
  Conference Proceedings, 2019.

\bibitem[Speer et~al.(2017)Speer, Chin, and Havasi]{speer2017conceptnet}
Robyn Speer, Joshua Chin, and Catherine Havasi.
\newblock {ConceptNet} 5.5: An open multilingual graph of general knowledge.
\newblock pages 4444--4451, 2017.
\newblock URL \url{http://aaai.org/ocs/index.php/AAAI/AAAI17/paper/view/14972}.

\bibitem[Vinyals et~al.(2019)Vinyals, Babuschkin, Czarnecki, Mathieu, Dudzik,
  Chung, Choi, Powell, Ewalds, Georgiev, Oh, Horgan, Kroiss, Danihelka, Huang,
  Sifre, Cai, Agapiou, Jaderberg, and Silver]{vinyals2019alphastar}
Oriol Vinyals, Igor Babuschkin, Wojciech Czarnecki, Michaël Mathieu, Andrew
  Dudzik, Junyoung Chung, David Choi, Richard Powell, Timo Ewalds, Petko
  Georgiev, Junhyuk Oh, Dan Horgan, Manuel Kroiss, Ivo Danihelka, Aja Huang,
  Laurent Sifre, Trevor Cai, John Agapiou, Max Jaderberg, and David Silver.
\newblock Grandmaster level in starcraft ii using multi-agent reinforcement
  learning.
\newblock \emph{Nature}, 575, 11 2019.
\newblock \doi{10.1038/s41586-019-1724-z}.

\end{thebibliography}

\end{document}